\documentclass{article}
\usepackage{spconf,amsmath,graphicx}

\usepackage{enumitem}
\setlist{nosep, leftmargin=14pt}

\usepackage{mwe} 

\title{The impact of training dataset size and ensemble inference strategies on head and neck auto-segmentation}
\name{Edward G. A. Henderson, Marcel van Herk, Eliana M. Vasquez Osorio}
\address{Division of Cancer Sciences, The University of Manchester,\\Oxford Rd, Manchester, M13 9PL}

\newcommand\blfootnote[1]{%
  \begingroup
  \renewcommand\thefootnote{}\footnote{#1}%
  \addtocounter{footnote}{-1}%
  \endgroup
}

\begin{document}
%
\maketitle
\begin{abstract}
Convolutional neural networks (CNNs) are increasingly being used to automate segmentation of organs-at-risk in radiotherapy. Since large sets of highly curated data are scarce, we investigated how much data is required to train accurate and robust head and neck auto-segmentation models. For this, an established 3D CNN was trained from scratch with different sized datasets (25-1000 scans) to segment the brainstem, parotid glands and spinal cord in CTs. Additionally, we evaluated multiple ensemble techniques to improve the performance of these models. The segmentations improved with training set size up to 250 scans and the ensemble methods significantly improved performance for all organs. The impact of the ensemble methods was most notable in the smallest datasets, demonstrating their potential for use in cases where large training datasets are difficult to obtain.
\end{abstract}
\begin{keywords}
auto-segmentation, radiotherapy, ensemble methods, data-efficient deep learning
\end{keywords}

\blfootnote{\textbf{Accepted in 20th IEEE International Symposium on Biomedical Imaging (ISBI 2023)}}
\vspace{-7mm}
\section{Introduction}
\label{sec:intro}
Half of people will get cancer in their lifetime and many of these will receive radiotherapy in their treatment\cite{Ahmad2015}. Radiotherapy patients receive a computed tomography (CT) scan which is used to plan their treatment. Accurate 3D segmentation of healthy organs close to the tumour (also known as organs-at-risk, OARs) in the CT is critical to plan the best treatment, focusing radiation onto the tumour (the target) and sparing OARs. Manual segmentation of OARs by clinicians is slow and prone to variability\cite{Brouwer2012}, so convolutional neural networks (CNNs) are now increasingly being used to automatically generate segmentations.

It is generally well known that deep learning models trained on larger datasets will generalise better to unseen data. However, in radiotherapy, large sets of high-quality training data are scarce and annotating large 3D CT scans is time-consuming. Therefore, it is important to understand how many examples are required to train a CNN model for 3D OAR auto-segmentation that is accurate and robust. In this study, we evaluated this for head and neck (HN) auto-segmentation. We additionally evaluated several ensemble techniques, combining the predictions from multiple trained models, which could boost segmentation performance.
\vspace{-3mm}
\section{Material and Methods}
\vspace{-1mm}
\label{sec:mmat}
For this study we gathered, from a single institution, 1215 planning CT scans with clinical OAR segmentations for the brainstem, parotid glands and the cervical section of the spinal cord. 
We reserved 215 scans as an unseen test set which left 1000 scans for model training.

\begin{figure}[ht]
\centering
\centerline{\includegraphics[width=\linewidth]{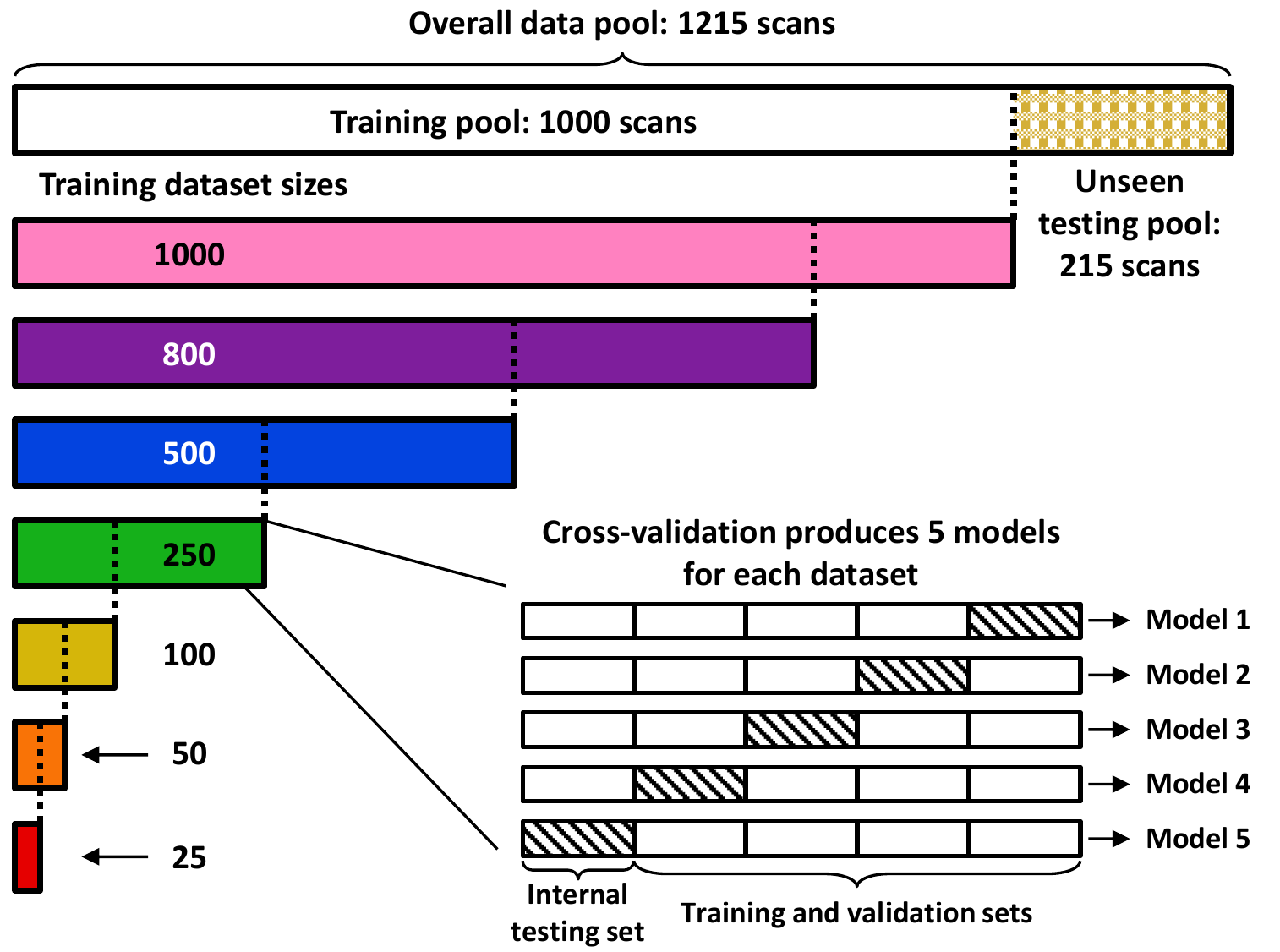}}
\caption{A schematic showing how the data pool of 1215 HN CT scans was divided. 215 scans are held out separately for testing (gold check pattern). The remaining 1000 are subdivided into multiple training dataset sizes ranging from 25 to 1000 scans. A 5-fold cross-validation is completed for each dataset size. The internal test sets (hatched) were used to determine the best-performing cross-validation model.}
\label{fig:data_pool}
\end{figure}

We trained an established HN auto-segmentation CNN\cite{Henderson2022} to segment these OARs from scratch with random subsets of either 25, 50, 100, 250, 500, 800 and 1000 images. A 5-fold cross-validation was performed on each subset, splitting the data from each to produce 5 unique parameter sets (or models), Fig~\ref{fig:data_pool}. The models produced from each dataset size were evaluated on the unseen test set using 2 inference strategies.

\vspace{-4mm}
\subsection{Inference strategies}
\label{ssec:inf_strats}
First, the best model (BM) from each 5-fold cross-validation was selected based on the segmentation performance in the internal test set for each fold (shown hatched in Fig~\ref{fig:data_pool}). Models were ranked based on the median values of the bi-directional mean distance-to-agreement (mDTA) and 95th percentile Hausdorff distance (HD95) metrics.

Next, we tested four different ensembling techniques to combine the predictions of all 5 models at each dataset size:

\begin{enumerate}
    \item \textbf{Summation of logits} - the raw predicted probabilities from each model were summed prior to taking the argmax to generate a segmentation mask. 
    \item \textbf{Summation of Softmax} - the probabilities from each model were passed through a Softmax activation function prior to being summed and then argmax applied to generate a segmentation mask. 
    \item \textbf{Majority vote} - argmax was applied directly to generate a predicted segmentation mask for each model and then the most popular class for each voxel was computed.
    \item \textbf{STAPLE} - argmax was applied to generate a segmentation mask for each model and then the Simultaneous Truth and Performance Level Estimation (STAPLE) algorithm \cite{Warfield2004} was used to generate a consensus segmentation mask. 
\end{enumerate}

Both BM and each of the ensemble inference strategies were tested in the unseen testing pool of 215 scans, and the segmentation quality again assessed using mDTA and HD95. These two metrics are complementary as mDTA evaluates the overall quality of the segmentation and HD95 highlights large errors. We used Wilcoxon signed-rank tests to compare each of the ensemble approaches to the BM inference to determine which had superior performance.

We ranked all the ensemble strategies based on the level of improvement using a simple points-based system, allocating a number of points for the level of significance of the improvement suggested by the Wilcoxon signed-rank test. We gave points as follows: \textbf{5} for p $<0.000005$; \textbf{4} for p $<0.00005$; \textbf{3} for p $<0.0005$; \textbf{2} for p $<0.005$; \textbf{1} for p $<0.05$; \textbf{0} for p $>0.05$ (not significant). We report the total number of points for each ensemble method and metric, across all OARs and dataset sizes. While rudimentary, this ranking system summarizes which ensemble techniques perform best. 

\subsection{Implementation details}
The CNN model used in this study was a 3D UNet with residual connections. Full implementation details can be found in \cite{Henderson2022} and identical training protocols were used for each model. 
All CNN models were implemented in PyTorch 1.10.1 and training was performed using a single NVidia GeForce RTX 3090 GPU with 24GB of memory.

\section{Results}
\label{sec:results}
\begin{figure*}[t!]
\centering
\centerline{\includegraphics[width=18cm]{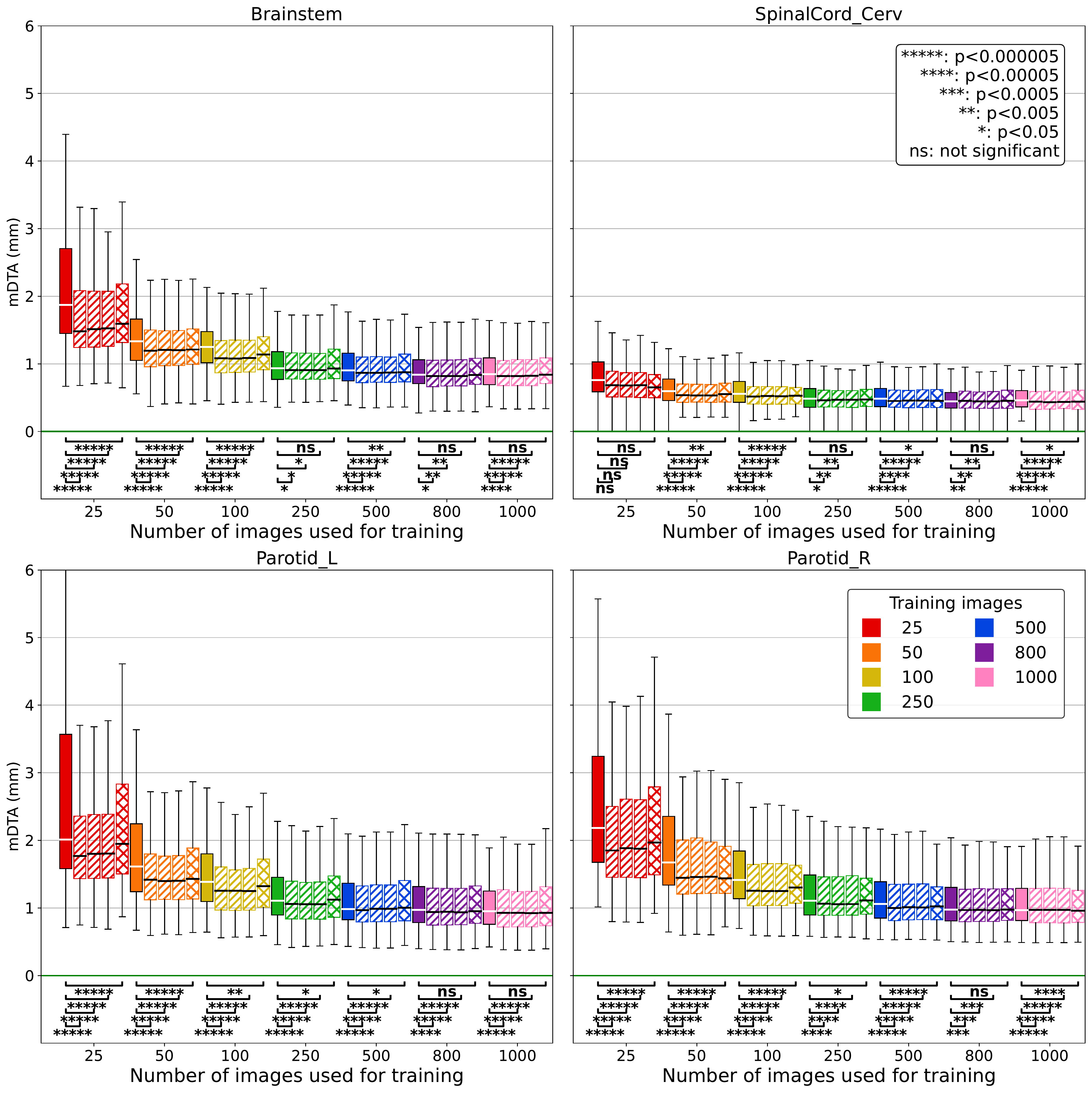}}
\caption{Boxplots showing the mDTA results in the unseen test set of 215 scans for each OAR. The different colours indicate dataset size. The solid boxes are for inference with the single best model and the hatched boxes are using the ensemble methods in the order, left-to-right, they are described in section~\ref{ssec:inf_strats}. The cross-hatched boxes show the results of the STAPLE ensemble method. The level of significance of the differences are shown with asterisks at the base of each plot. Lower is better for mDTA.}
\label{fig:res}
\end{figure*}

Figure~\ref{fig:res} shows the mDTA results of models for every dataset size. The results for different inference styles are shown in neighbouring boxes. The solid-coloured boxes show results of BM while the hatched boxes are the results with different ensemble techniques in the order described in section~\ref{ssec:inf_strats}. As expected, the auto-segmentation performance steadily improves with dataset size. Above 250 scans we observed very little improvement in segmentation performance.

Each of the ensemble techniques improved the segmentation performance when compared to inference with the BM alone, and ensemble techniques 1-3 are significantly better for all dataset sizes for the mDTA and HD95 metric.

The use of the STAPLE algorithm results in worse performance compared to the other ensemble techniques for the mDTA metric, occasionally performing slightly worse than the single best model inference (e.g. for 250 \& 800 images).

Interestingly, STAPLE performs similarly to the other combination techniques when measured via the HD95 (results not included due to space restrictions). 
This suggests that while the STAPLE algorithm is producing lesser quality segmentations, it is not producing gross errors.  Indeed when looking at the mean of differences in the volume of the ensemble methods segmentations and the gold standard, the STAPLE method systematically overestimated segmentations (Table~\ref{tab:vol}).

\vspace{-2mm}
\begin{table}[ht!]
\caption{Mean volume differences to the gold standard across all OARs and datasets. Closer to zero is better.}\vspace{1mm}
\centering
\begin{tabular}{lc}
 & mean volume diff. (cm$^3$) \\ \hline
Summation of logits & 0.21 \\
Summation of Softmax & 0.12 \\
Majority vote & 0.11 \\
STAPLE & 1.55 \\ \hline
\end{tabular}
\label{tab:vol}
\end{table}

In Table~\ref{tab:res} we show the summary rankings of each of the ensemble techniques. These rankings indicate that the summation of Softmax and majority vote techniques performed best in our experiments. On the other extreme, the STAPLE algorithm ranked the lowest, indicating poorer performance when combining segmentations. However, it is worth noting that the STAPLE algorithm still yielded improved segmentation performance when compared to simply using the prediction of the single best model from the cross-validation (BM).  
\vspace{-3mm}
\begin{table}[h]
\caption{Summary rankings for the ensemble methods. Higher is better. }\vspace{1mm}
\centering
\begin{tabular}{lccc}
 & mDTA $\uparrow$ & HD95 $\uparrow$ & Total $\uparrow$ \\ \hline
Summation of logits & 120 & 81 & 201 \\
Summation of Softmax & 123 & 82 & 205 \\
Majority vote & 124 & 81 & 205 \\
STAPLE & 70 & 76 & 146 \\ \hline
\end{tabular}
\label{tab:res}
\end{table}

\section{Discussion}
\label{sec:discussion}

In this study we have explored the question of how much data is needed to successfully train a CNN for 3D CT auto-segmentation.  For our case, segmenting HN OARs, we observed that improvements in performance were negligible when training with datasets larger than 250 clinical samples.

Additionally, we tested different ensemble inference strategies, including STAPLE, a common algorithm used in other applications (e.g., atlas-based segmentation, \cite{Vrtovec2020}).
We found that the ensemble strategies of either summing up the Softmax maps or using majority vote performed significantly better. Moreover, the segmentation performance boost of the ensemble techniques were most significant as the number of training examples decreased, showing the most dramatic improvement for the smallest datasets.

Fang et al. similarly explored the impact of training sample size on HN deep-learning auto-segmentation models\cite{Fang2021}. However, their segmentation results were based solely on the Dice similarity coefficient which, whilst popular, is volume biased, insensitive to fine details and can hide clinically relevant differences between structure boundaries\cite{Sharp2014}. Their study also used a 2D model which may have different performance characteristics with dataset size than our 3D model. Ren et al. used ensembling to combine predictions from multiple imaging modalities (CT, MRI and PET) for tumour segmentation\cite{Ren2021}. They used an average of Softmax probabilities, similar to our best performing ensemble approach.

Whilst we have shown that segmentation performance scales with dataset size and can be improved with ensemble techniques, it is evident this is not the entire story. The CNN architecture used in this study was originally designed for limited data and demonstrated good performance when trained with just 34 CTs with highly-consistent, arbitrated gold-standard segmentations\cite{Henderson2022, nikolov2018}. However, in our study, the models were trained on the unedited clinical contours which are susceptible to observer variation. Therefore, it is apparent there are further gains in performance to be attained by cultivating a highly-consistent training dataset.

We have shown that ensemble techniques can be used to improve the performance of auto-segmentation models, with an effect that is most noticeable for models trained with smaller datasets. Ensemble methods could be particularly effective in scenarios where data is scarce, for example in cases of rare anatomies or for structures that are difficult to segment. In such cases, ensemble methods offer improved robustness and accuracy of automatic segmentations despite a limited number of training examples.

\section{Conclusion}
\label{sec:conclusion}
We have demonstrated that ensemble techniques significantly improve auto-segmentation performance of healthy organs in the head and neck region. Furthermore, ensemble inference strategies are an effective technique to improve segmentations, especially in cases where larger validated training datasets are difficult to obtain.


\section{Compliance with ethical standards}
\label{sec:ethics}

This project used retrospective and anonymised patient images, and was approved by the UK Computer Aided Theragnostics Research Database Management Committee (North West - Haydock Research Ethics Committee reference number: 21/NW/0347).

\section{Acknowledgments}
\label{sec:acknowledgments}
Marcel van Herk was supported by NIHR Manchester Biomedical Research Centre. This work was also supported by Cancer Research UK via funding to the Cancer Research Manchester Centre [C147/A25254] and by Cancer Research UK RadNet Manchester [C1994/A28701].

\bibliographystyle{IEEEbib}
\bibliography{strings,refs}

\end{document}